\documentclass[runningheads]{llncs}
\usepackage{graphicx}
\usepackage{amsmath}
\usepackage{algorithm}
\usepackage{algorithmic}
\usepackage{array}
\usepackage{cite}
\usepackage[caption=false]{subfig}
\usepackage{url}
\usepackage{multirow}
\usepackage{adjustbox}
\begin{document}
%
\title{Stacked Autoencoder Based Feature Extraction and Superpixel Generation for Multifrequency PolSAR Image Classification}
\titlerunning{Autoencoder Based Multifrequency PolSAR Image Classification}
%
\author{Tushar Gadhiya, Sumanth Tangirala, and Anil K. Roy}
\authorrunning{Tushar Gadhiya, Sumanth Tangirala, and Anil K. Roy}
%
\institute{Dhirubhai Ambani Institute of Information and Communication Technology}
%
\maketitle              
\begin{abstract}
In this paper we are proposing classification algorithm for multifrequency Polarimetric Synthetic Aperture Radar (PolSAR) image. Using PolSAR decomposition algorithms 33 features are extracted from each frequency band of the given image. Then, a two-layer autoencoder is used to reduce the dimensionality of input feature vector while retaining useful features of the input. This reduced dimensional feature vector is then applied to generate superpixels using simple linear iterative clustering (SLIC) algorithm. Next, a robust feature representation is constructed using both pixel as well as superpixel information. Finally, softmax classifier is used to perform classification task. The advantage of using superpixels is that it preserves spatial information between neighbouring PolSAR pixels and therefore minimizes the effect of speckle noise during classification. Experiments have been conducted on Flevoland dataset and the proposed method was found to be superior to other methods available in the literature.

\end{abstract}
\section{Introduction}
Synthetic Aperture Radar (SAR) has been popularized in recent years as a technique that captures high resolution microwave images of the earth surface. With SAR technique an image can be taken regardless of weather conditions or time of the day unlike optical sensors. The other major reason to use SAR is its operability over multiple frequency bands, viz, from the X band to P band. The penetrability of the L and the P bands allows SAR to capture data from even below ground level. In case of PolSAR it draws information of the target in four polarization states which makes it an information rich technique. These are some of the reasons that establish the superiority of SAR over optical data capturing techniques.
\par One of the very early approaches of classification of multifrequency PolSAR data was attempted using DNN (Dynamic Neural Network) \cite{multidyna}. Deep learning based multiplayer autoencoder network was also proposed \cite{tensor}. It uses Kronecker product of eigenvalues of coherency matrix to combine multiple bands information. Recently, an Optimized Wishart Network (OWN) for classification of multifrequency PolSAR data is reported \cite{own}.
\par Superpixel algorithm in conjunction with deep neural networks has gained popularity for capturing spatial information of a PolSAR image. Hou \textit{et al.} presented \cite{spixelae} a way of using Pauli decomposition of PolSAR image to generate superpixels and autoencoder to extract features from the coherency matrix of each PolSAR pixel. Prediction of the network was then used to run a KNN (k nearest neighbours) algorithm in each superpixel to determine the class of the complete superpixel. Guo \textit{et al.} introduced a method to apply Fuzzy clustering algorithm over PolSAR images to generate superpixels \cite{fuzzysp}. This method considered only those pixels which are similar to their neighbours, pixels which are in all probability badly conditioned were ignored. In another work \cite{dfic} Cloude decomposition features were used to generate superpixels and CNN (Convolutional Neural Network) to perform the classification. Adaptive nonlocal approach for extracting spatial information was also proposed \cite{anssa}. It uses stacked sparse autoencoder to extract robust features.
\par In this paper we propose a classification algorithm for multifrequency PolSAR images. For doing so, we extracted 33 features from each band of PolSAR image to construct a feature vector for each pixel. After that we used two-layer stacked autoencoder to reduce dimensionality of the input vector. The output of autoencoder which contained information of all features of all bands was used to generate superpixels. Next, each pixel along with its corresponding superpixel was used to construct robust feature vector. Finally softmax classifier was used to perform the classification task.
\par In this paper we propose a classification algorithm for multifrequency PolSAR images. This involves extraction of 33 features from each band which are then used by a two-layer stacked autoencoder to reduce dimensionality of the input vector. This is used to generate superpixels whose features are used to generate a feature vector including information regarding the pixel and the superpixel the pixel belongs to. The merging of this information ensures that the effect of any noise or error can be reduced on the pixels. After this, the softmax layer to used for classification.
\par This paper is organized as follows: Section II discusses the proposed network architecture; Section III explains the experiments conducted on the Flevoland dataset; Section IV concludes with our observations and the discussions based on the results.
\section{Proposed Methodology}
The raw data of a PolSAR image is a 2x2 scattering matrix where each element denotes the phase and magnitude information of the received waves in vertical and horizontal polarization after scattering of vertically and horizontally polarized waves as 4 complex numbers:
\[S=\begin{bmatrix}
S_{hh}&S_{hv}\\
S_{vh}&S_{vv}\\
\end{bmatrix},\]
\par Due to the reciprocity condition, it is assumed that \(S_{hv} = S_{vh}\). So we have 3 complex numbers where each complex number has 2 elements-real and imaginary. If a neural network were to be given the 6 elements as input, it wouldn't be able to compute the phase information since it considers the real and imaginary components as separate features. This brings the need to give crafted features to the network instead of raw data. We have created 33 dimensional feature vector extracted from one frequency band of a PolSAR image. It contains 6 features of coherency matrix \cite{cohf}, 3 features of Freeman decomposition \cite{freeman}, 4 features of Krogager decomposition \cite{kro}, 4 features of Yamaguchi decomposition \cite{yamaguchi2}, 9 features of Huynen decomposition \cite{huynen} and 7 features of Cloude decomposition \cite{cloudepot} as shown in Table \ref{features}. Hence combining information of all three bands we get 99 dimensional feature vector corresponding to each PolSAR pixel.
\par The Coherency Matrix contains the 6 elements from the scattering matrix. Freeman decomposition has 3 features which describe the power of single-bounce(odd-bounce), double-bounce and volume scattering of the transmitted waves. The Huynen decomposition aims to form a single scattering matrix to model the scattering mechanism of the surface. The Cloude decomposition was proposed in 1997 and it aims to model the surface scattering using the parameters such as Entropy, Anisotropy and other angles. The Krogager decomposition aims to factorize the scattering matrix as the combination of a sphere, a diplane and a helix. The Yamaguchi decomposition adds a Helix scattering component to the Freeman decomposition to model complicated manmade structures.
\begin{table}
	\renewcommand{\arraystretch}{1.2}
	\centering
	\caption{List of extracted features}
	\label{features}
	\begin{tabular}{p{5.5cm}|m{6.5cm}}
		\hline
		Features & Description\\
		\hline
		${|T_{13}|}/{\sqrt{T_{11}T_{33}}}$, ${|T_{23}|}/{\sqrt{T_{33}T_{22}}}$, ${T_{22}}/{S}$, ${T_{33}}/{S}$, $10log_{10}(S)$, ${|T_{12}|}/{\sqrt{T_{11}T_{22}}}$  & 6 features of Coherency matrix \cite{cohf} \\
		\hline
		$F_{dbl}$, $F_{odd}$, $F_{vol}$ & 3 features of Freeman decomposition \cite{freeman} \\
		\hline
		$K_d$, $K_h$, $K_s$, $K_t$ & 4 features of Krogager decomposition\cite{kro} \\
		\hline
		$P_{dbl}$, $P_{hlx}$, $P_{odd}$, $P_{vol}$ & 4 features of Yamaguchi decomposition \cite{yamaguchi2} \\
		\hline
		$A$, $B_0$, $B$, $C$, $D$, $E$, $F$, $G$, $H$ & 9 features of Huynen decomposition \cite{huynen} \\		
		\hline
		$\alpha$, $\beta$, $\delta$, $\gamma$, $\lambda$, Entropy, Anisotropy & 7 features of Cloude decomposition \cite{cloudepot} \\

		\hline
	\end{tabular}
\end{table}
\par As shown in Fig. \ref{network} our proposed network architecture contains three modules. First module is a two layer autoencoder network. The purpose of this module is to reduce the dimensionality of the input vector by learning efficient representation of combined PolSAR frequency bands information.
\par Let $\mathbf{X_i}$ be the input feature vector of $i^{th}$ PolSAR pixel. Let $\mathbf{W_{11}}$, $\mathbf{W_{12}}$, $\mathbf{b_{11}}$ and $\mathbf{b_{12}}$ be weights and biases of two encoder layers. A hidden representation of input feature vector $\mathbf{X_i}$ can be calculated as $\mathbf{h_i} = f(\mathbf{W_{12}}^Tf(\mathbf{W_{11}}^T\mathbf{X_i} + \mathbf{b_{11}})+\mathbf{b_{12}})$, where $f$ is a $tanh$ activation function. Let $\mathbf{W_{21}}$, $\mathbf{W_{22}}$, $\mathbf{b_{21}}$ and $\mathbf{b_{22}}$ be weights and biases of two decoder layers. A reconstructed input vector can be calculated as $\mathbf{X_i}^{'} = f(\mathbf{W_{22}}^Tf(\mathbf{W_{21}}^T\mathbf{h_i} + \mathbf{b_{21}})+\mathbf{b_{22}})$. We have used mean square error to train the autoencoder. The cost function of first module of the proposed network is given as follows:
\begin{equation}
\begin{split}
J_1 &= \beta (\mathbf{\| W_{11} \|^2_2 + \| W_{12} \|^2_2 + \| W_{21} \|^2_2 + \| W_{22} \|^2_2}) \\
 &+ \alpha \frac{1}{N} \sum_{i=1}^{N} \|\mathbf{X_i - X_i}^{'}\|^2
+\gamma \sum_{t=1}^{U_1} KL(\rho \| \rho_t).
\end{split}
\end{equation}
Here $\beta$ is a regularization parameter, $\alpha$ is the learning rate, $\gamma$ is the sparsity parameter, N is total number of training samples, $U_1$ is a size of reduced dimensional feature vector $\mathbf{h_i}$, $\rho$ is a sparsity parameter, $\rho_t$ is the average activation value of the $t^{th}$ hidden unit and $KL(\rho \| \rho_t)$ is a Kullback-Leibler divergence which encourages sparsity in the hidden representation $\mathbf{h_i}$. For our application we have set the value of $U_1=5$. Once the training of the first module is complete we disconnect the decoder layers.
\par Next, we feed entire PolSAR image as an input to this network to obtain hidden representations of all pixels of the PolSAR image. We will use this hidden representation of all pixels of the PolSAR image to generate superpixels. This process has two advantages: i) it contains feature information of all bands and ii) its dimensionality is substantially reduced in comparison to the input feature vector. To generate superpixels we have used algorithm similar to simple linear iterative clustering (SLIC) \cite{slic}. Instead of the RGB input image we will use hidden representation of PolSAR image obtained from using first module of the proposed network. Using SLIC we measure the distance between any two pixels which is given by Eq. \ref{sup}:
\begin{figure}[t]
	\centering
	\includegraphics[width=4.5in]{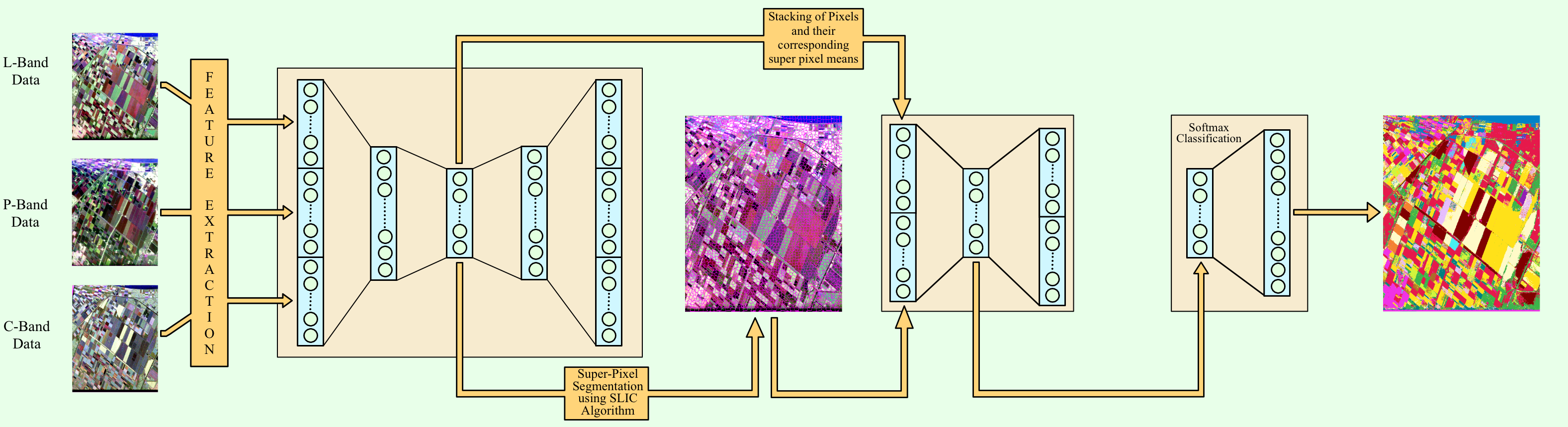}
	\caption{Network architecture distributed in 3 modules showing that the extracted features of all the 3 bands are fed into a stacked autoencoder for dimensionality reduction followed by generation of superpixels. Now each pixel is stacked with its corresponding superpixel mean for robust feature generation. We used Softmax classifier to generate classification map.}
	\label{network}
\end{figure}
\begin{equation} \label{sup}
D = \frac{m}{s}D_s + D_h,
\end{equation}
where $D_s = \sqrt{(x_i-x_j)^2+(y_i - y_j)^2}$ and $D_h = \lVert \mathbf{h_i}-\mathbf{h_j} \rVert_2$. Here $m$ is a parameter controlling the relative weight between $D_s$ and $D_h$ and $s$ is a size of search space \cite{slic}. $(x_i, y_i)$ and $(x_j, y_j)$ are the positions of $i^{th}$ and $j^{th}$ PolSAR pixels on Euclidean plane. This distance measure was finally used for superpixel generation. 
\par The second module of our proposed architecture combines each pixel and corresponding superpixel information to construct robust feature vector. This is done by letting $\mathbf{S_j}$ be the $j^{th}$ superpixel and $\mathbf{h_i} \in \mathbf{S_j}$. Let $\mathbf{c_j}$ be the cluster center of $\mathbf{S_j}$. To extract robust feature using both pixel and superpixel information input for second autoenocder $\mathbf{H_i}$ can be constructed as $\mathbf{H_i} = [\mathbf{h_i}; \mathbf{c_j}]$ \cite{anssa}. Since the dimensionality of the hidden representation $h_i$ is low, a single layer autoencoder is sufficient for an effective reconstruction. Let $\mathbf{r_i} = f(\mathbf{W_{13}}^T\mathbf{H_i} + \mathbf{b_{13}})$ be the activation value obtained at hidden layer of second autoencoder. Let $\mathbf{H_i^{'}} = f(\mathbf{W_{23}}^T\mathbf{r_i} + \mathbf{b_{23}})$ be the reconstructed input. The cost function for the second autoencoder can now readily be described by:
\begin{equation}
\begin{split}
J_2 = \beta (\| \mathbf{W_{13}} \|^2_2 + \| \mathbf{W_{23}} \|^2_2)
+ \lambda \frac{1}{N} \sum_{i=1}^{N} \|\mathbf{H_i} - \mathbf{H_i}^{'}\|^2
&+ \alpha \sum_{t=1}^{U_2} KL(\rho \| \rho_t).
\end{split}
\end{equation}
Here $U_2$ is the size of feature vector $\mathbf{r_i}$. Once the training of the second module of the autoencoder is complete we again disconnect the decoder layer. Output of the second module of autoencoder contains both pixel and superpixel information. Finally we use softmax classifier to obtain predicted probability distribution.
\section{Experiments}
With this theoretical model we conduct the following experiments. We start with the details of dataset chosen for our experiments. After that we analyze the performance of the proposed network for different band combinations. Finally we will perform the analysis of performance for different feature decompositions. The complete experiment on the proposed network architecture was implemented using Python 3.6 and it was executed on a 1.60 GHz machine with 8GB RAM for all the experiments.
\par
\begin{figure}
	\centering
	\subfloat[]{\includegraphics[width=1in]{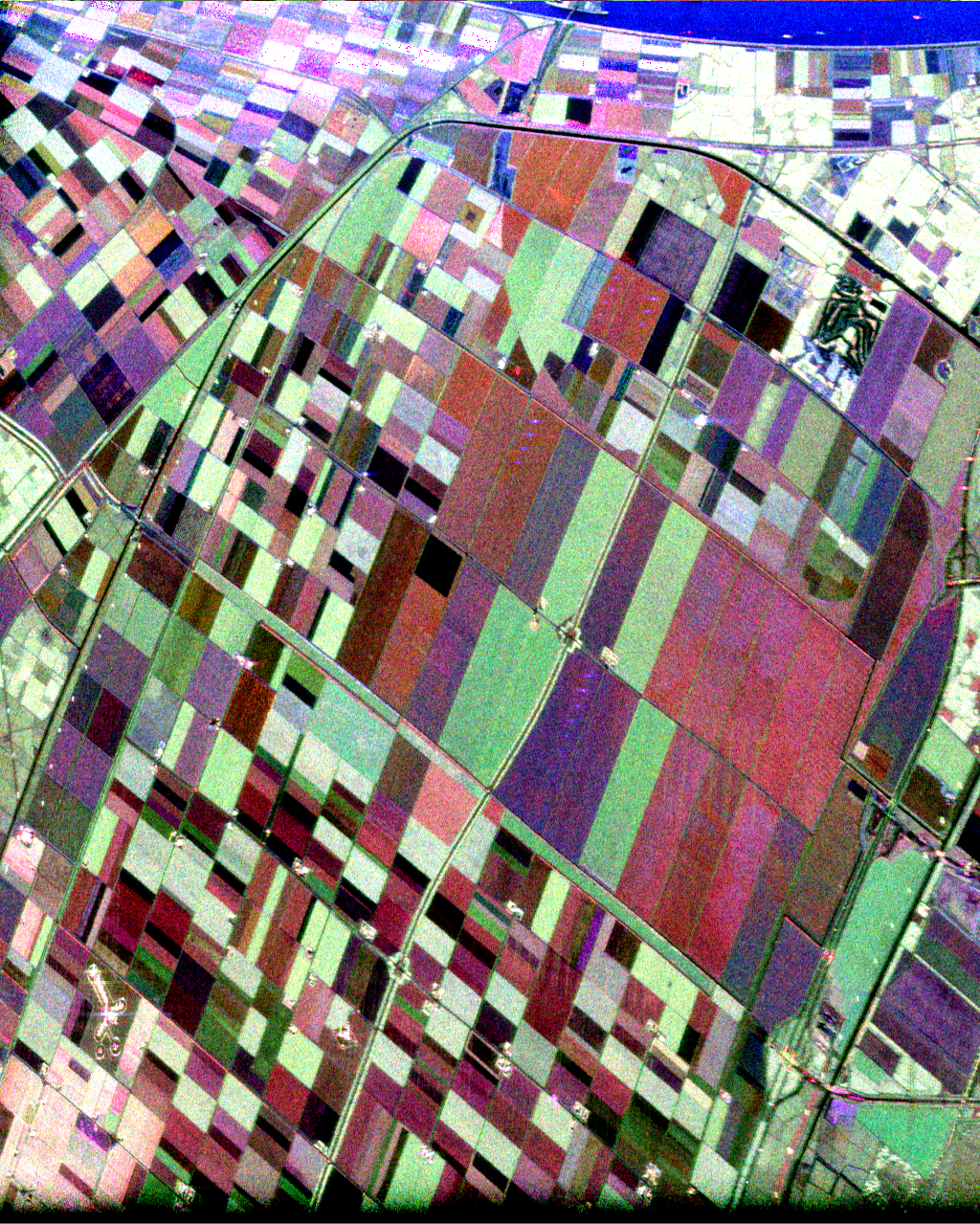}\label{fig:pauill}}~
	\subfloat[]{\includegraphics[width=1in]{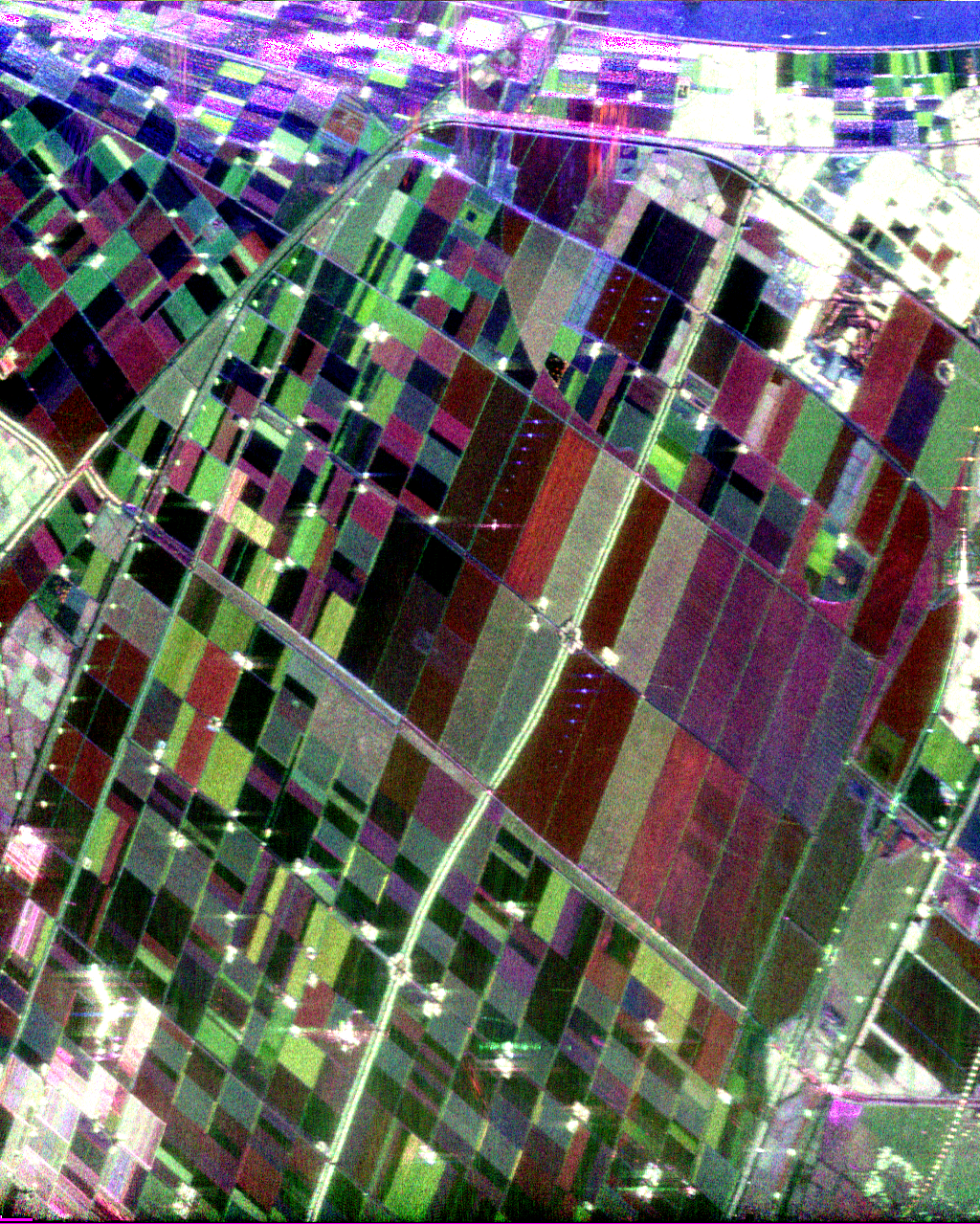}\label{fig:pauilp}}~
	\subfloat[]{\includegraphics[width=1in]{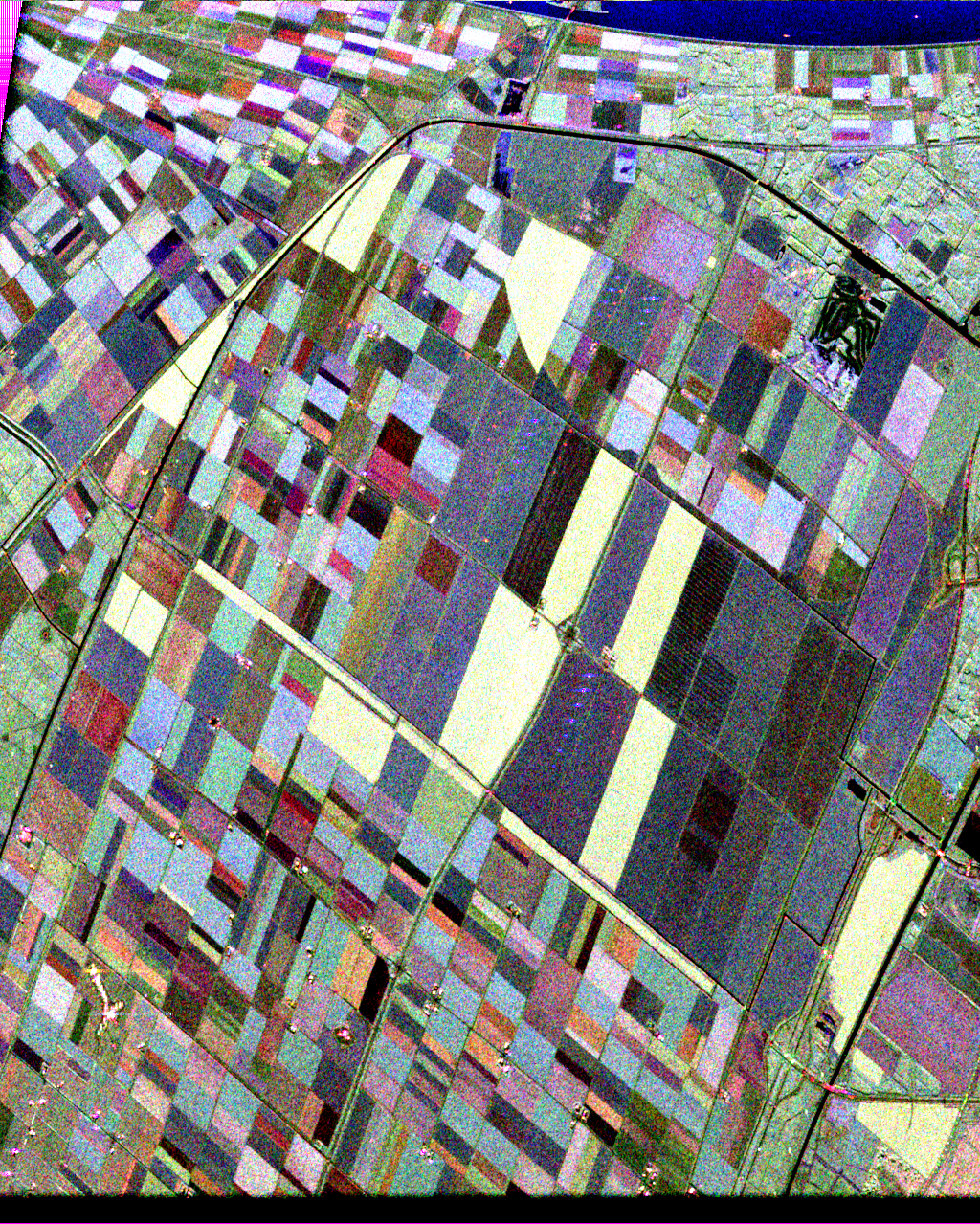}\label{fig:pauilc}}~
	\subfloat[]{\includegraphics[width=1in]{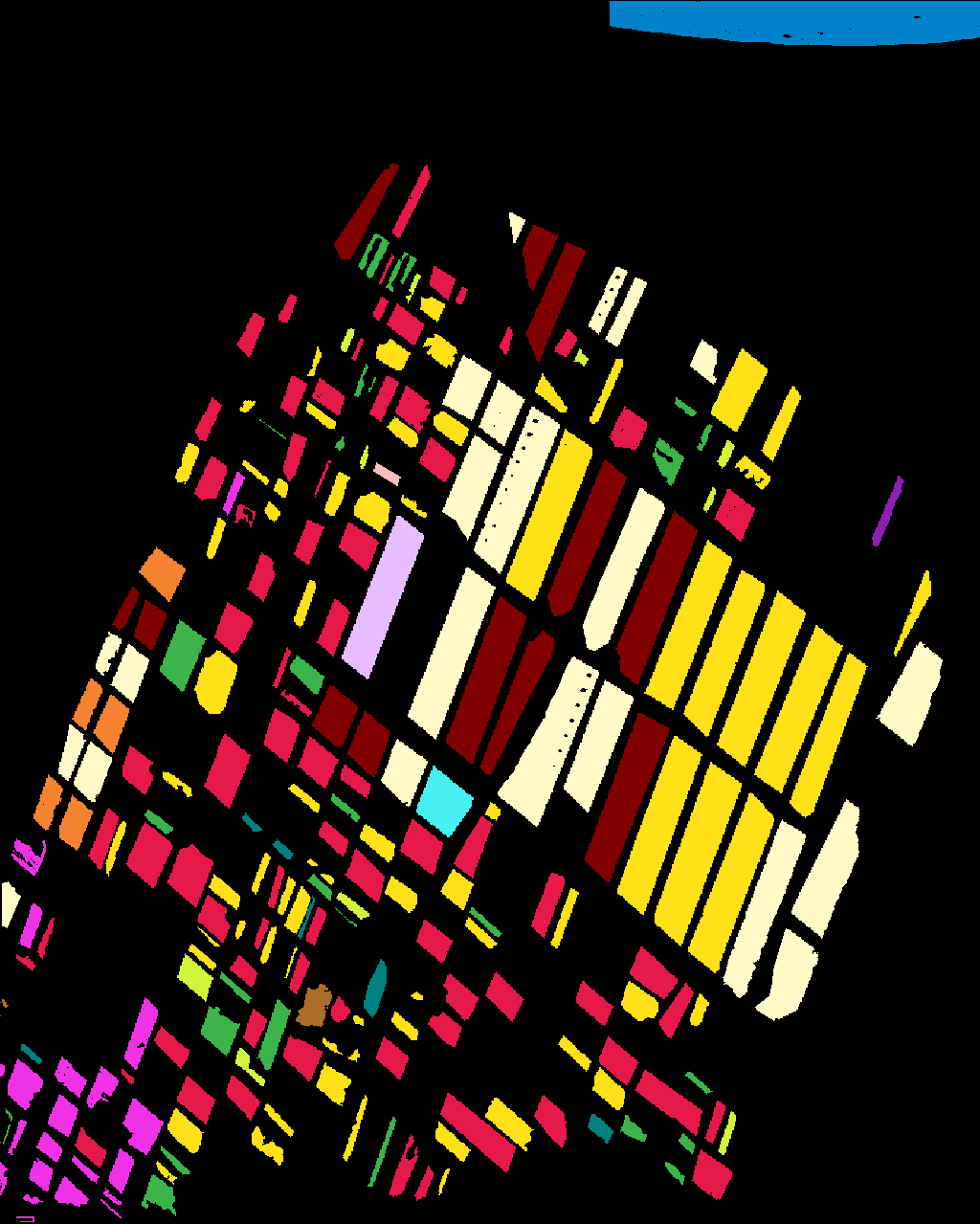}\label{fig:gt}} \\
	\includegraphics[width=4.5in]{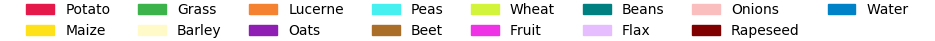}
	\caption{Pauli decomposition of (a) L band, (b) P band and (c) C band Flevoland dataset \cite{airsar}. (d) Ground truth map of Flevoland dataset}
	\label{dataset}
\end{figure}

 Experiments have been conducted on a dataset of Flevoland \cite{airsar}, an agricultural tract in the Netherlands, whose data was captured by the NASA/Jet Propulsion Laboratory on 15 June, 1991. This dataset has often been viewed as the benchmark dataset for PolSAR applications. The intensities after Pauli decomposition of the dataset have been used to form an RGB image as shown in Fig. \ref{dataset}(a-c). The ground truth of the data set shown in Fig. \ref{dataset}(d) identifies a total of 15 classes of land cover.
\begin{figure*}
	\centering
	\subfloat[]{\includegraphics[width=1in]{gt.png}\label{fig:gtt}}~
	\subfloat[]{\includegraphics[width=1in]{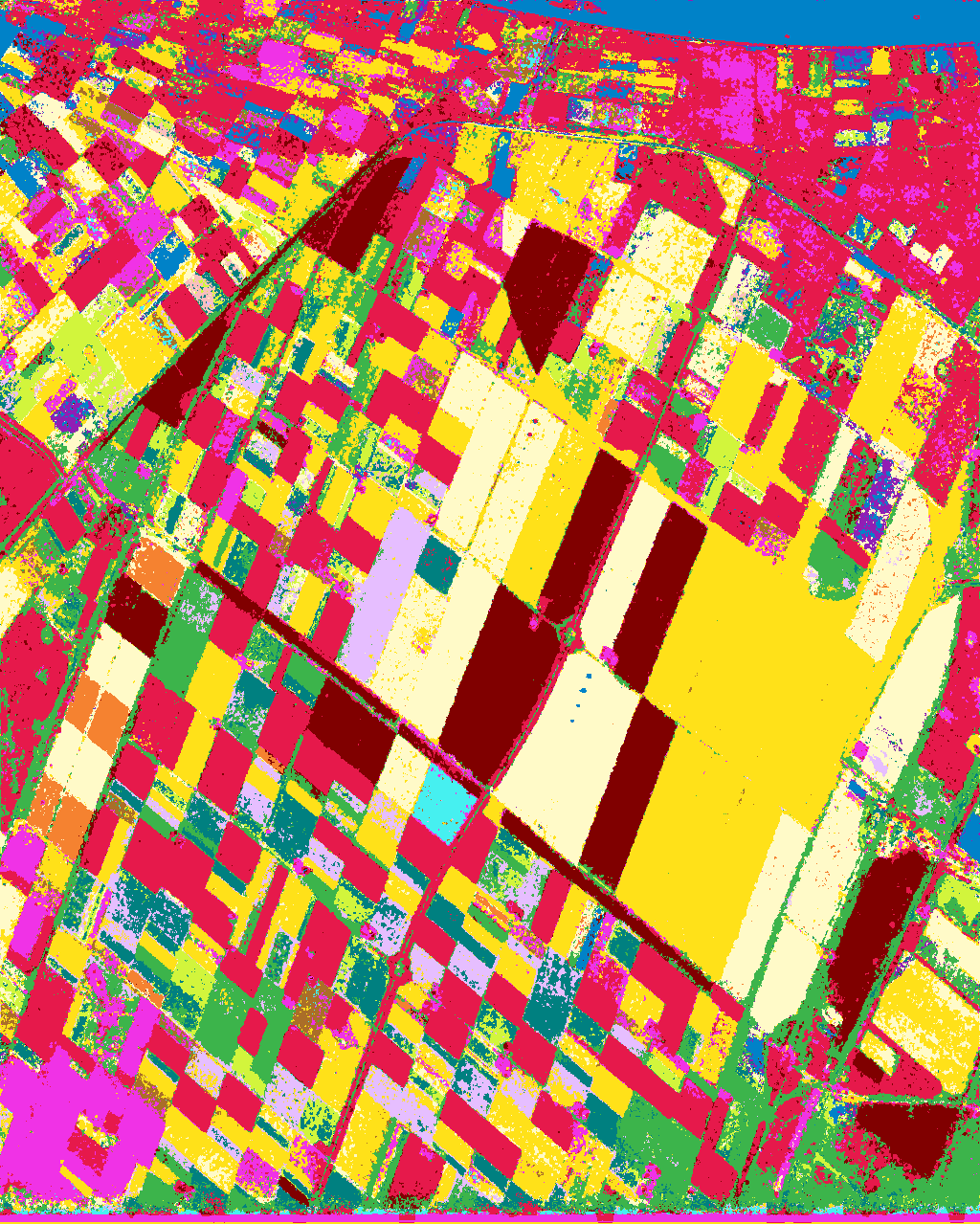}\label{fig:l}}~
	\subfloat[]{\includegraphics[width=1in]{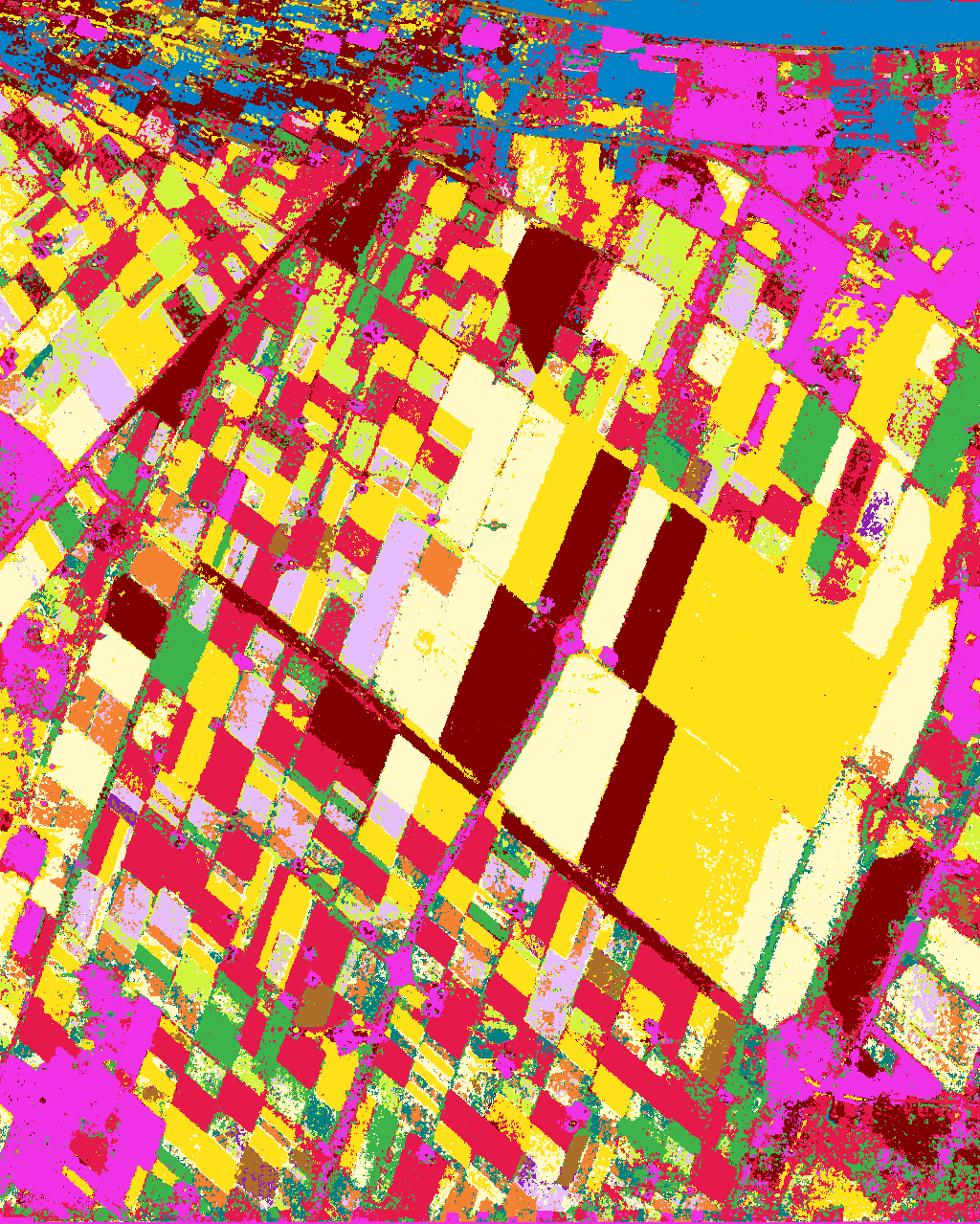}\label{fig:p}}~
	\subfloat[]{\includegraphics[width=1in]{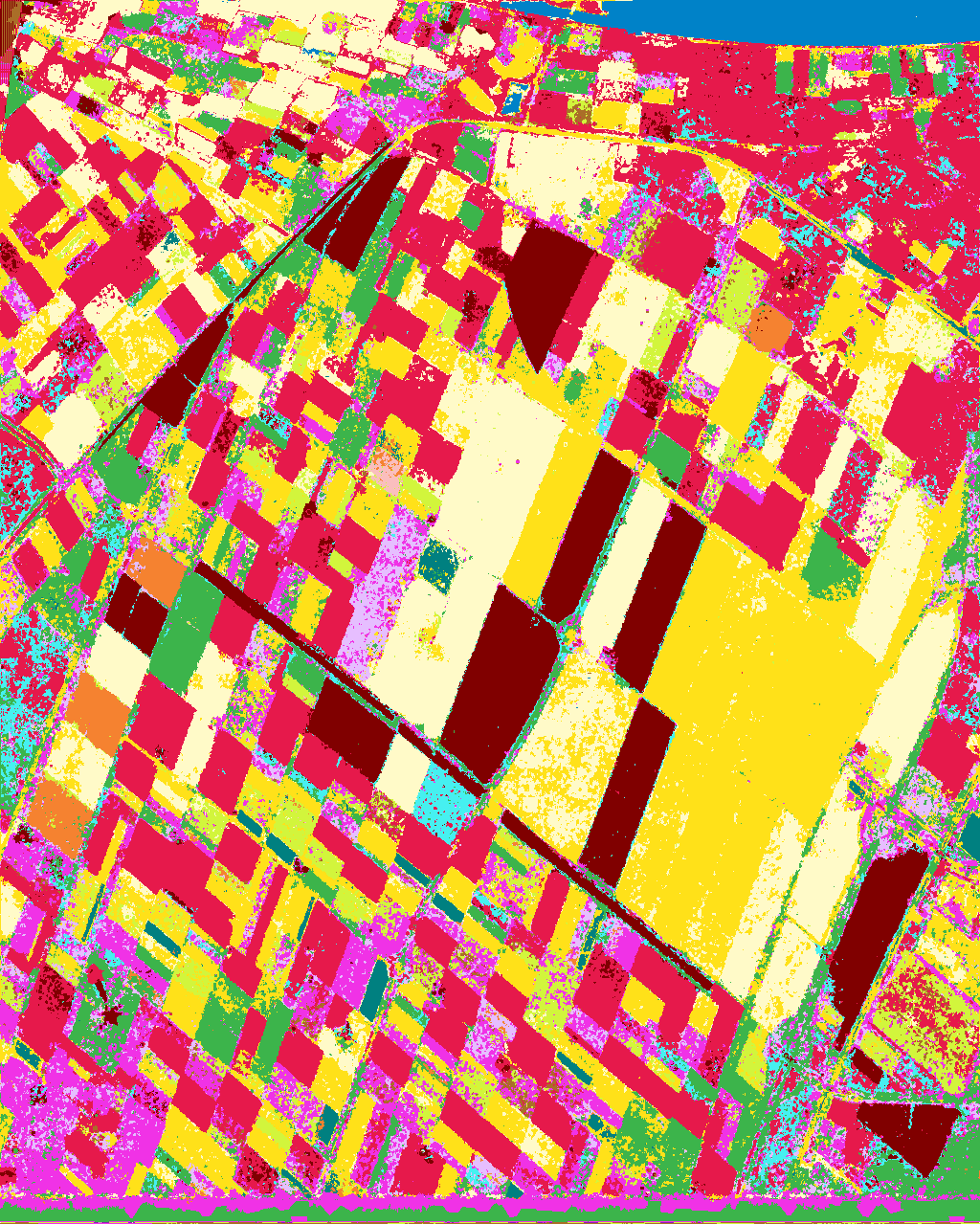}\label{fig:c}}~ \\
	\subfloat[]{\includegraphics[width=1in]{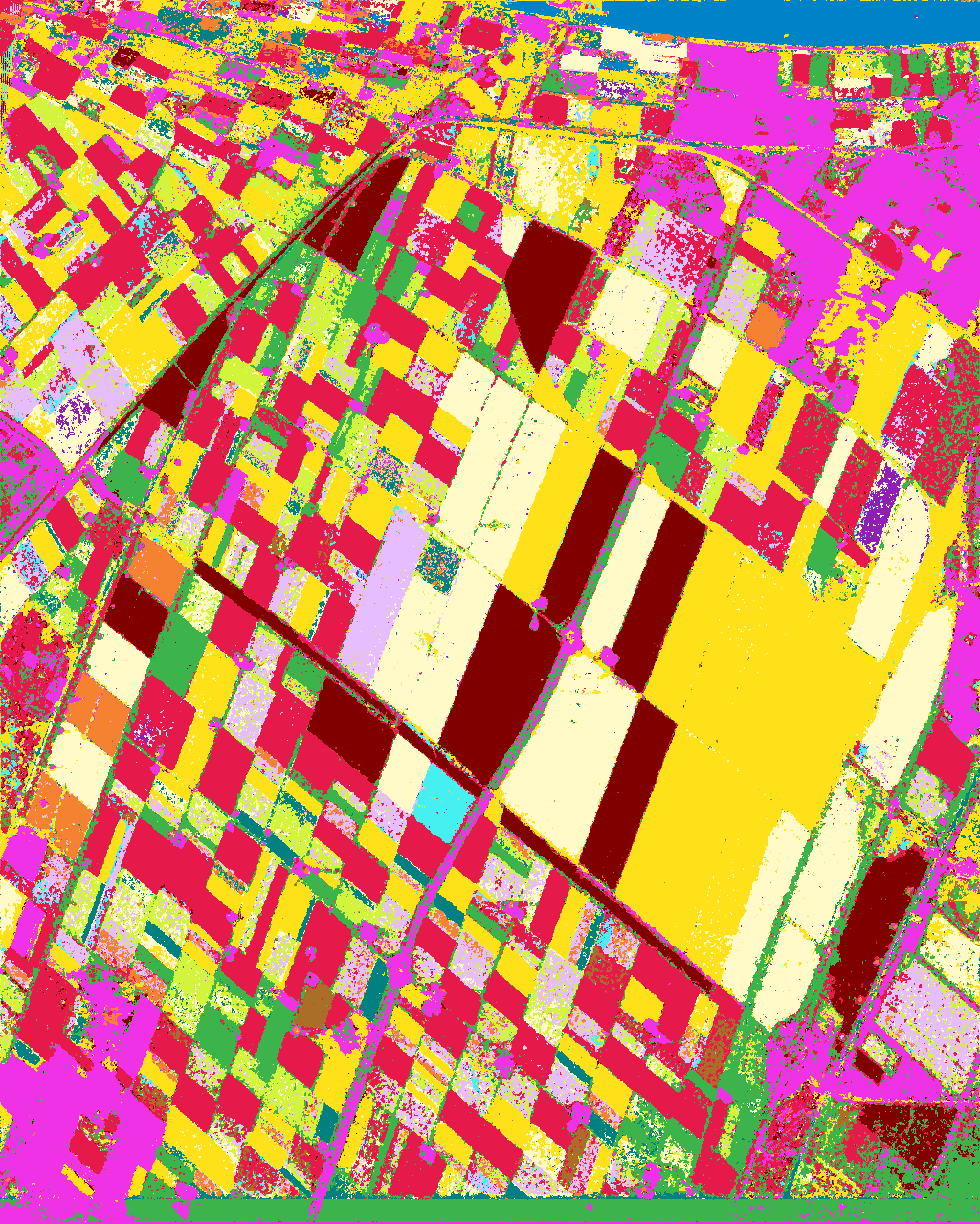}\label{fig:pc}}~
	\subfloat[]{\includegraphics[width=1in]{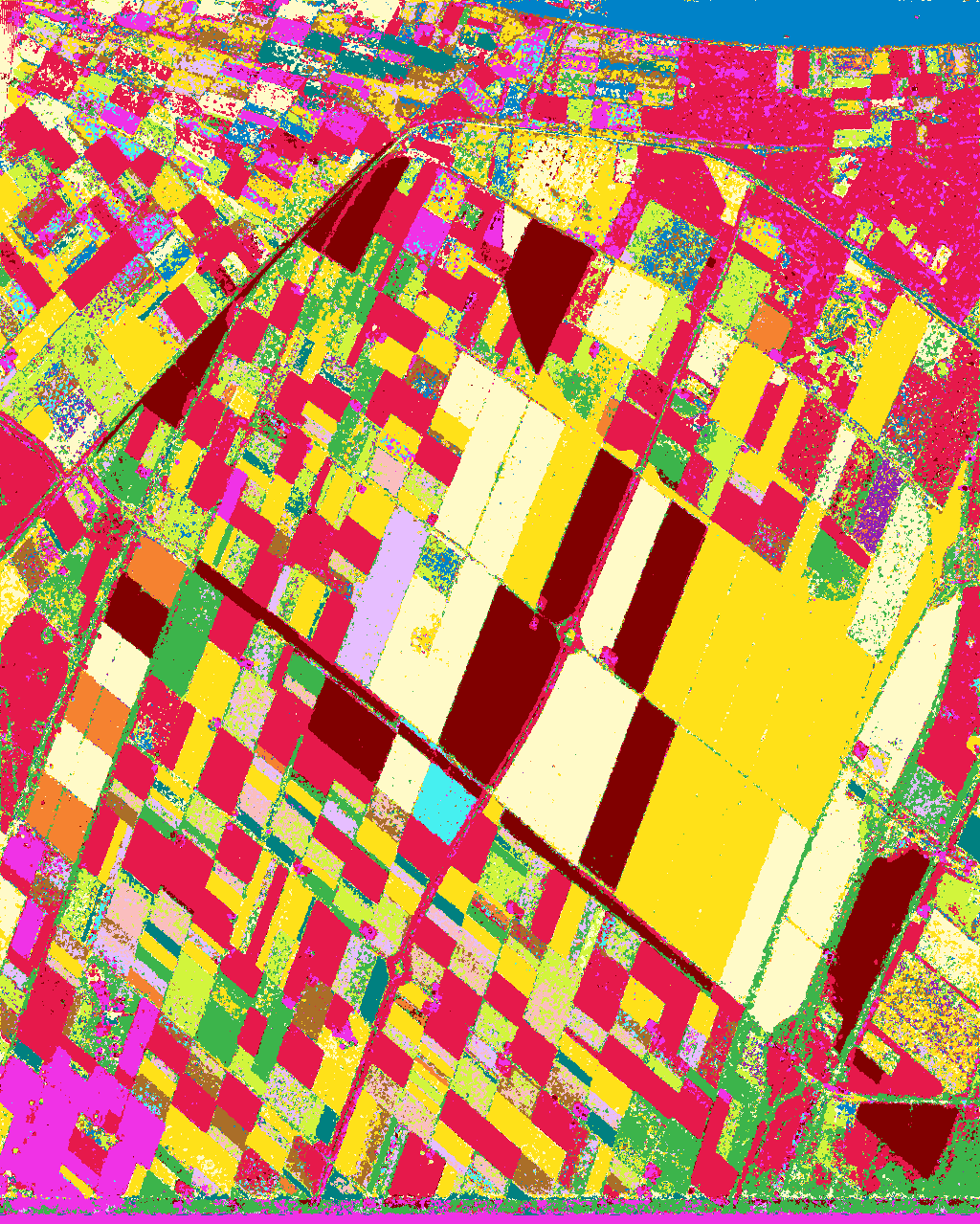}\label{fig:lc}}~
	\subfloat[]{\includegraphics[width=1in]{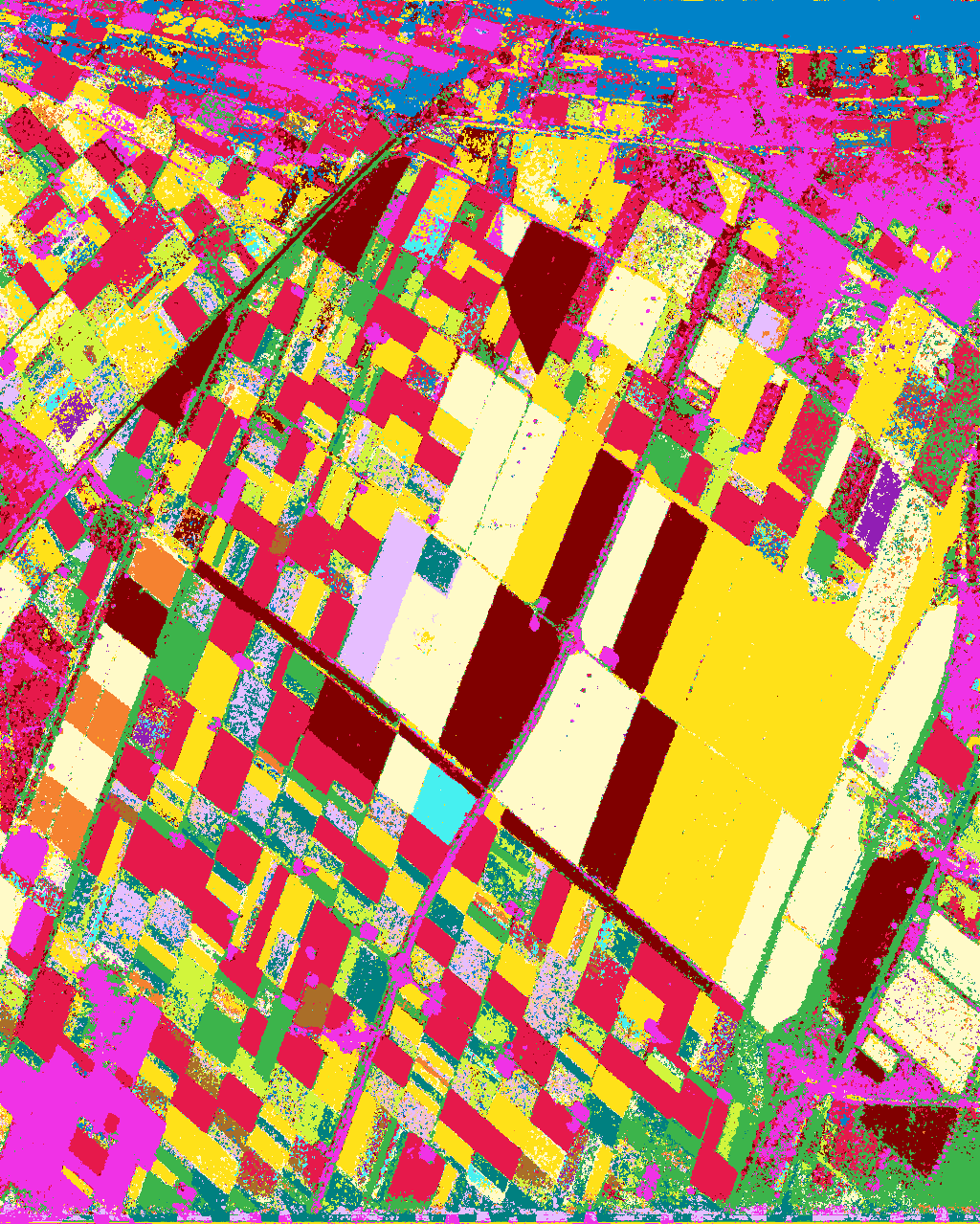}\label{fig:lp}}~
	\subfloat[]{\includegraphics[width=1in]{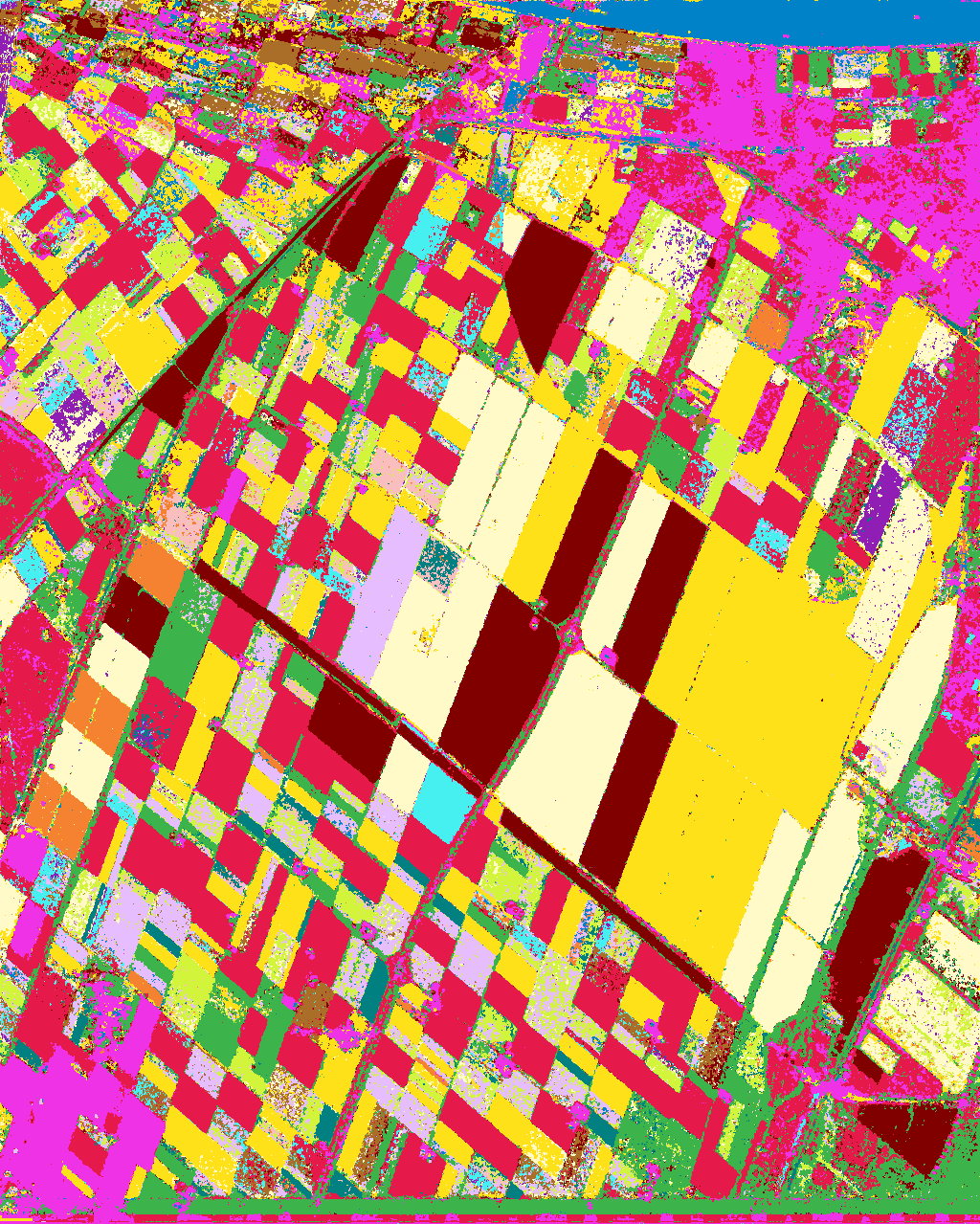}\label{fig:lpc}} \\
	\includegraphics[width=4.5in]{lagend.png}
	\caption{(a) Ground truth map, Classification map of Flevoland dataset obtained using (b) L band, (c) P band, (d) C band, (e) P and C band , (f) L and C band, (g) L and P band and (h) L, P and C band.}
	\label{maps}
\end{figure*}
\begin{table}
	\centering
	\caption{Class-wise accuracies for different band combinations}
	\label{band}
	\renewcommand{\arraystretch}{1.2}
	\begin{tabular}{ p{2cm}p{1.1cm}p{1.1cm}p{1.1cm}p{1.1cm}p{1.1cm}p{1.1cm}p{1.1cm}  }
		\hline
		Class/Bands		& L        & P      & C      & LP     & LC     & PC     & LPC  \\
		\hline
		Potato   & 0.9912 & 0.9636 & 0.9890 & 0.9977 & 0.9961 & 0.9931 & 0.9975 \\
		Maize    & 0.9859 & 0.9745 & 0.9292 & 0.9963 & 0.9932 & 0.9917 & 0.9979 \\
		Grass    & 0.9339 & 0.8282 & 0.8585 & 0.9741 & 0.9457 & 0.9608 & 0.9773 \\
		Barley   & 0.9757 & 0.9758 & 0.8848 & 0.9960 & 0.9930 & 0.9857 & 0.9979 \\
		Lucerne  & 0.9506 & 0.8148 & 0.9573 & 0.9798 & 0.9879 & 0.9638 & 0.9869 \\
		Oats     & 0.4641 & 0.1771 & 0.0000 & 0.8916 & 0.6275 & 0.7466 & 0.8718 \\
		Peas     & 0.9642 & 0.0024 & 0.8975 & 0.9952 & 0.9827 & 0.9807 & 0.9952 \\
		Beet     & 0.8024 & 0.9490 & 0.1667 & 0.9608 & 0.9545 & 0.9627 & 0.9772 \\
		Wheat    & 0.8756 & 0.6940 & 0.4792 & 0.9375 & 0.9328 & 0.8633 & 0.9538 \\
		Fruit    & 0.9505 & 0.9819 & 0.7591 & 0.9931 & 0.9674 & 0.9906 & 0.9944 \\
		Beans    & 0.8954 & 0.4990 & 0.8784 & 0.9410 & 0.9767 & 0.9534 & 0.9850 \\
		Flax     & 0.9856 & 0.9409 & 0.6749 & 0.9984 & 0.9860 & 0.9718 & 0.9966 \\
		Onions   & 0.2353 & 0.0000 & 0.8893 & 0.4706 & 0.9273 & 0.2907 & 0.9275 \\
		Rapeseed & 0.9952 & 0.9951 & 0.9982 & 0.9986 & 0.9994 & 0.9988 & 0.9994 \\
		Water    & 0.9979 & 0.9964 & 0.9997 & 0.9975 & 0.9983 & 0.9983 & 0.9995 \\
		\hline
		OA       & \textbf{0.9784} & \textbf{0.9508} & \textbf{0.9211} & \textbf{0.9938} & \textbf{0.9900} & \textbf{0.9869} & \textbf{0.9959} \\
		\hline
	\end{tabular}
\end{table}
\par We have evaluated the accuracy of each class with respect to all possible combinations of the data acquired in the three frequency bands. Please note that all the Overall Accuracy mentioned in the paper is the precision obtained. From Table \ref{band} it can be observed that while the network with just the C band information recognizes Onions and Lucerne with high accuracy, it fails in the case of classes such as Beet or Oats. While in the case of the C band, the network fails to recognize Wheat accurately. It can be noted that the lack of information to recognize Wheat is compensated by the help of the L band. In the case of the P band, the network fails to accurately identify Onions and Peas but correctly identifies the majority of the Rapeseed and Fruit. It can also be observed that the L band performs better individually than the C or the P bands.
\par A total of six PolSAR decomposition methods have been applied to the PolSAR data and their features have been given as input to the proposed network. The significance of each decomposition technique is evident from the Table \ref{feature_acc}. It can be seen that the Krogager and Freeman decompositions fail to provide enough information to recognize Beet and Onions. On the other hand Cloude decomposition and Huynen decomposition provide sufficient information respectively.
\begin{table}
	\centering
	\caption{Class-wise accuracies for different features}
	\label{feature_acc}
		\begin{tabular}{ p{1.3cm}p{1.6cm}p{1.6cm}p{1.6cm} p{1.6cm}p{1.6cm}p{1.6cm}}
			\hline
			/ & Yamaguchi & Coherency & Krogager & Freeman & Hynen & Cloude \\
			\hline
			Potato  & 0.9820 & 0.9860 & 0.9879 & 0.9721 & 0.9933 & 0.9941 \\
			Maize   & 0.9672 & 0.9696 & 0.9880 & 0.9465 & 0.9904 & 0.9884 \\
			Grass   & 0.9243 & 0.8898 & 0.8490 & 0.6757 & 0.9256 & 0.9555 \\
			Barley  & 0.9653 & 0.9516 & 0.9814 & 0.9554 & 0.9920 & 0.9846 \\
			Lucerne & 0.6690 & 0.8568 & 0.4815 & 0.4820 & 0.9695 & 0.9522 \\
			Oats    & 0.0626 & 0.0031 & 0.2168 & 0.0000 & 0.3420 & 0.7252 \\
			Peas    & 0.9779 & 0.8686 & 0.7440 & 0.5245 & 0.9546 & 0.9904 \\
			Beet    & 0.1658 & 0.6794 & 0.0000 & 0.0000 & 0.8834 & 0.9244 \\
			Wheat   & 0.1460 & 0.7210 & 0.6841 & 0.0000 & 0.8916 & 0.9428 \\
			Fruit   & 0.9618 & 0.9661 & 0.9827 & 0.9027 & 0.9790 & 0.9874 \\
			Beans   & 0.7831 & 0.9265 & 0.8758 & 0.0160 & 0.9767 & 0.9167 \\
			Flax    & 0.9346 & 0.9636 & 0.7162 & 0.7470 & 0.9790 & 0.9688 \\
			Onions  & 0.0000 & 0.2561 & 0.0000 & 0.0000 & 0.8720 & 0.1246 \\
			R.seed  & 0.9921 & 0.9949 & 0.9976 & 0.9786 & 0.9976 & 0.9986 \\
			Water   & 0.9889 & 0.9957 & 0.9958 & 0.9926 & 0.9981 & 0.9968 \\
			\hline
			\textbf{OA} & \textbf{0.9509} & \textbf{0.9609} & \textbf{0.9562} & \textbf{0.9105} & \textbf{0.9859} & \textbf{0.9856}\\
			\hline
		\end{tabular}
\end{table}
\par The proposed method for classification of multifrequency PolSAR image is also compared with other methods available in the literature. Comparison of overall accuracies is reported in Table \ref{acc_comp}. ANN \cite{tensor} and OWN \cite{own} have used small subset of Flevoland dataset containing 7 classes. On the other hand  Stein-SRC \cite{rem} used ground truth with 14 classes. For parity in comparison with our results we have calculated the overall accuracy of the proposed method using ground truth of 7, 14 as well as 15 classes. We observe from  Table \ref{acc_comp} that the proposed method outperforms all the three methods \cite{own, tensor, rem}.
\begin{table}
	\centering
	\caption{Classification's overall accuracy comparison}
	\label{acc_comp}
	\begin{tabular}{lcc}
		\hline
		Method & Number of classes & Accuracy \\
		\hline
		ANN \cite{tensor} & 7 & 98.23 \\
		OWN \cite{own} & 7 & 98.56 \\
		Stein-SRC \cite{rem} & 14 & 99.00 \\
		\hline
		Proposed Method & 7& 99.93\\
		Proposed Method & 14& 99.69\\
		Proposed Method & 15& 99.59\\
		\hline
	\end{tabular}
\end{table}
\section{Conclusion}
In this paper a classification network for multifrequency PolSAR data is proposed. Proposed network involves three modules. First module reduces the dimensionality of the input feature vector. Output of the first module has been used to generate superpixels. The second module constructs robust feature vector using each pixel and its corresponding superpixel information. Finally the last module of the proposed network conducts classification using softmax classifier. It is observed that combining multiple frequency bands information improves overall classification accuracy. We validated our proposed network on Flevoland dataset resulted in 99.59\% overall classification accuracy. Experimental result shows that this proposed network outperforms other reported methods available in the literature.
\bibliographystyle{ieeetr}
\bibliography{bib}
\end{document}